\documentclass{article}

    \PassOptionsToPackage{numbers}{natbib}

\usepackage[final]{neurips_wrl2020}

\usepackage[utf8]{inputenc} 
\usepackage[T1]{fontenc}    
\usepackage{hyperref}       
\usepackage{url}            
\usepackage{booktabs}       
\usepackage{amsfonts}       
\usepackage{nicefrac}       
\usepackage{microtype}      

\usepackage{amsmath}
\usepackage{amssymb}
\usepackage{multirow}
\usepackage{tabularx}
\usepackage{cuted}
\usepackage{siunitx}
\usepackage{wrapfig}
\usepackage[font=tiny,labelfont=bf]{caption}
\usepackage[export]{adjustbox}
\usepackage{color}
\usepackage{siunitx}
\usepackage{placeins}
\usepackage{amsmath}
\usepackage{amsfonts}
\usepackage{amssymb}
\usepackage[normalem]{ulem}
\usepackage[disable]{todonotes}
\usepackage{algorithm}
\usepackage{algorithmic}
\usepackage{float}
\usepackage{subfigure}
\usepackage{layouts}
\usepackage{listings}
\usepackage{enumitem}
\usepackage{printlen}
\usepackage{array}
\DeclareMathOperator*{\argmax}{argmax}

\newcommand{\snu}{\vspace{-0.1em}}

\title{Towards Exploiting Geometry and Time for Fast Off-Distribution Adaptation in Multi-Task Robot Learning}

\author{
    K.R. Zentner\thanks{Equal Contribution} \\
    \texttt{kzentner@usc.edu} \\
    \And
    Ryan Julian\footnotemark[1] \\
    \texttt{rjulian@usc.edu} \\
    \And
    Ujjwal Puri \\
    \texttt{ujjwalpu@usc.edu} \\
    \And
    Yulun Zhang \\
    \texttt{yulunzha@usc.edu} \\
    \And
    Gaurav Sukhatme \\
    \texttt{gaurav@usc.edu} \\
    University of Southern California \\
    Los Angeles, CA 90089 \\
}

\begin{document}

\maketitle

\begin{abstract}
We explore possible methods for multi-task transfer learning which seek to exploit the shared physical structure of robotics tasks.  
Specifically, we train policies for a base set of pre-training tasks, then experiment with adapting to new off-distribution tasks, using simple architectural approaches for re-using these policies as black-box priors. These approaches include learning an alignment of either the observation space or action space from a base to a target task to exploit rigid body structure, and methods for learning a time-domain switching policy across base tasks which solves the target task, to exploit temporal coherence.
We find that combining low-complexity target policy classes, base policies as black-box priors, and simple optimization algorithms allows us to acquire new tasks outside the base task distribution, using small amounts of offline training data.
\end{abstract}

\section{Introduction}\snu
Real world robotics tasks all share the rich and predictable structure imposed by the laws of physics and the nature of the physical world.
The breakout success of deep learning approaches to domains such as computer vision, natural language processing, and recommender systems was precipitated by the design of model architectures which exploit the structure of the data in these domains, such as convolutional, auto-regressive, and graph neural networks respectively.
Despite these strong precedents, research in deep reinforcement learning (RL) and imitation learning (IL) for robotics has seen comparatively few attempts to design robotics-specific architectures transfer learning methods which exploit the structure of robotics tasks. We believe that multi-task robot learning in particular is likely to benefit from transfer methods which exploit large amounts of shared physical and temporal structure between tasks, because this structure is very likely to exist between tasks performed by a single robot design which is asked to perform many tasks in just one or a few environments.

\section{Problem Setting}\snu
We consider a multi-task reinforcement learning (RL) or imitation learning (IL) setting, defined by a possibly-unbounded discrete space of tasks $\mathcal{T}$. Each task $\tau \in \mathcal{T}$ is an infinite-horizon Markov decision process (MDP) defined by the tuple $(\mathcal{S}, \mathcal{A}, p, r_\tau)$. Motivated by our application to robotics, we presume all tasks in $\mathcal{T}$ share a single continuous state space $\mathcal{S}$, continuous action space $\mathcal{A}$, and state transition dynamics $p(s'|s,a)$, and so tasks are differentiated only by their reward functions $r_\tau(s,a)$. Our goal is to eventually learn policies $\pi_\tau(a|s)$ for each of $\tau \in \mathcal{T}$ which maximizes the expected total discounted return across all tasks $\tau \in \mathcal{T}$.

Importantly, we do not presume that the learner ever has access to all tasks in $\mathcal{T}$ at once, or even a representative sample thereof. Instead, we divide the lifetime of the learner into two phases. First pre-training, in which the learner is given access to a biased subset of tasks $\mathcal{T}_{\text{base}}$ and allowed to learn a near-optimal policy $\pi_\tau(a|s)$ for each task in $\mathcal{T}_{\text{base}}$, for a total of $|\mathcal{T}_{\text{base}}| = N_{\text{base}}$ \textit{base policies}. Second, adaptation, in which the learner is given access to data source $\mathcal{D}_{\text{target}}$ of trajectories for a target task $\tau_{\text{target}} \notin \mathcal{T}_{\text{base}}$ and base policies $\pi^{\tau \in \mathcal{T}_{\text{base}}}(a|s)$, and must quickly acquire a high-performance \textit{target policy} $\pi^{\text{target}}$ for the target task. As $\mathcal{T}_{\text{base}}$ is presumed to be a biased data-set with respect to $\mathcal{T}$, this is an off-distribution multi-task learning problem.

In this work, we presume $\mathcal{D}_{\text{target}}$ is static, small, and composed only of successful trajectories from an expert, so herein we discuss a few-shot imitation learning variant of this problem.

\section{Target Policy Classes for Structural Adaptation}\snu
In this section, we enumerate a set of low-dimensional target policy classes for fast adaptation between base and target tasks which share similar dynamical structure. In the next section, we measure the performance of these model classes under few-shot adaptation using a simulated 2D car-driving environment with non-trivial dynamics.

\subsection{Observation Alignment}\snu
Observation alignment uses a target policy class that contains a single source policy. This target policy class applies a transformation $T_\theta(s)$ to the observation before passing it to the base policy to generate actions. In our experiments we use an affine (linear transformation plus bias) transformation of the observation to simplify optimization, and to exploit simple geometric priors such as rigid transformation. As the optimization process is efficient, we choose a base policy by training a $T^\tau_\theta(a)$ for all $\tau \in \mathcal{T}_{\text{base}}$ and using the lowest loss member of the ensemble for the final target policy.

\begin{equation}
    \pi_\theta^{\text{target}}(a|s) = \pi^{\text{base}}(a|T_\theta(s))
\end{equation}

\subsection{Action (Re-)Alignment}\snu
Like observation alignment, action alignment learns a low-dimensional affine transformation function, but instead transforming the state input of a base policy, this function $T_\theta(a)$ transforms the action output. Like observation alignment, we train an ensemble of these target policies for a target task, and choose the one with the lowest loss.

\begin{equation}
    \pi_\theta^\text{target}(a|s) = \pi^{\text{base}}(T_\theta(a)|s)
\end{equation}

We find that naive action alignment performs poorly, because the final output layer of $\pi^{\text{base}}$ often destroys necessary information.
Action re-alignment instead uses all but the last layer of the base policy (hereafter referred to as $\pi^{\text{base}}_{-1}$) to encoder the current state into a latent encoding $h$.
Action re-alignment then uses a learned transformation function $T_\theta(h)$ to transform that encoding into an action on the target task.

\begin{equation}
    \pi_\theta^\text{target}(a|s) = \pi^{\text{base}}_{-1}(T_\theta(h)|s)
\end{equation}

\subsection{Time-Domain Switching and Mixing}\snu

\subsubsection{Mixing (Soft Switching)}\snu
In order to select actions in the target task, soft switching policy computes a state-conditioned weighting function over the base policies, $W_\theta(s,\tau)$.
It then computes an action distribution as a mixture of the base policies action distributions using that weighting.
\begin{equation}
    \pi_\theta^{\text{target}}(a|s) = \sum_{\tau \in \mathcal{T}_{\text{base}}} W_\theta(s, \tau)\pi^{\text{base},\tau}(a|s)
\end{equation}

\subsubsection{Hard Switching}\snu
As in soft switching, this target policy class computes a state-conditioned weighting function $W_\theta(s,\tau)$ over the base policies. Unlike soft switching, the hard switching policy uses actions from only a single base policy at any given time step. To discourage switching too often, the hard switching policy enforces hysteresis during sampling: it maintains a state, $h$ of the the most recently selected base policy, and continues to use that base policy until another base policy has a weight which is $\epsilon$ larger than the current base policy. 

\begin{equation}
    \pi_\theta^{\text{target}}(a|s, h) = \pi^{\text{base},h'}(a|s) 
\end{equation}
\begin{equation*}
    \quad\text{where}\quad 
    h' = 
    \begin{cases}
        h & \text{if}\ W_\theta(s,h) + \epsilon > W_\theta(s,\tau)\ \forall \tau \in \mathcal{\tau}_\text{base} \\
        \argmax_\tau W_\theta(s, \tau) & \text{otherwise}
    \end{cases}
\end{equation*}

\section{Experiments}\snu

\subsection{Training $T_\theta(s)$, $T_\theta(a)$, and $T_\theta(h)$}\snu
For this work, we have limited ourselves to affine $T$ transformations (i.e. $T(x) = Ax + b$ and $\theta = (A,b)$), which allows us to exploit a geometric prior common in robotics: that states and actions typically represent the position or velocity of rigid bodies in SE(3), or other grounded physical quantities which can be aligned between tasks using an affine transformation. 

\begin{figure}
    \centering
    \begin{minipage}[t]{0.45\columnwidth}
        \centering
        \includegraphics[width=\columnwidth,valign=t]{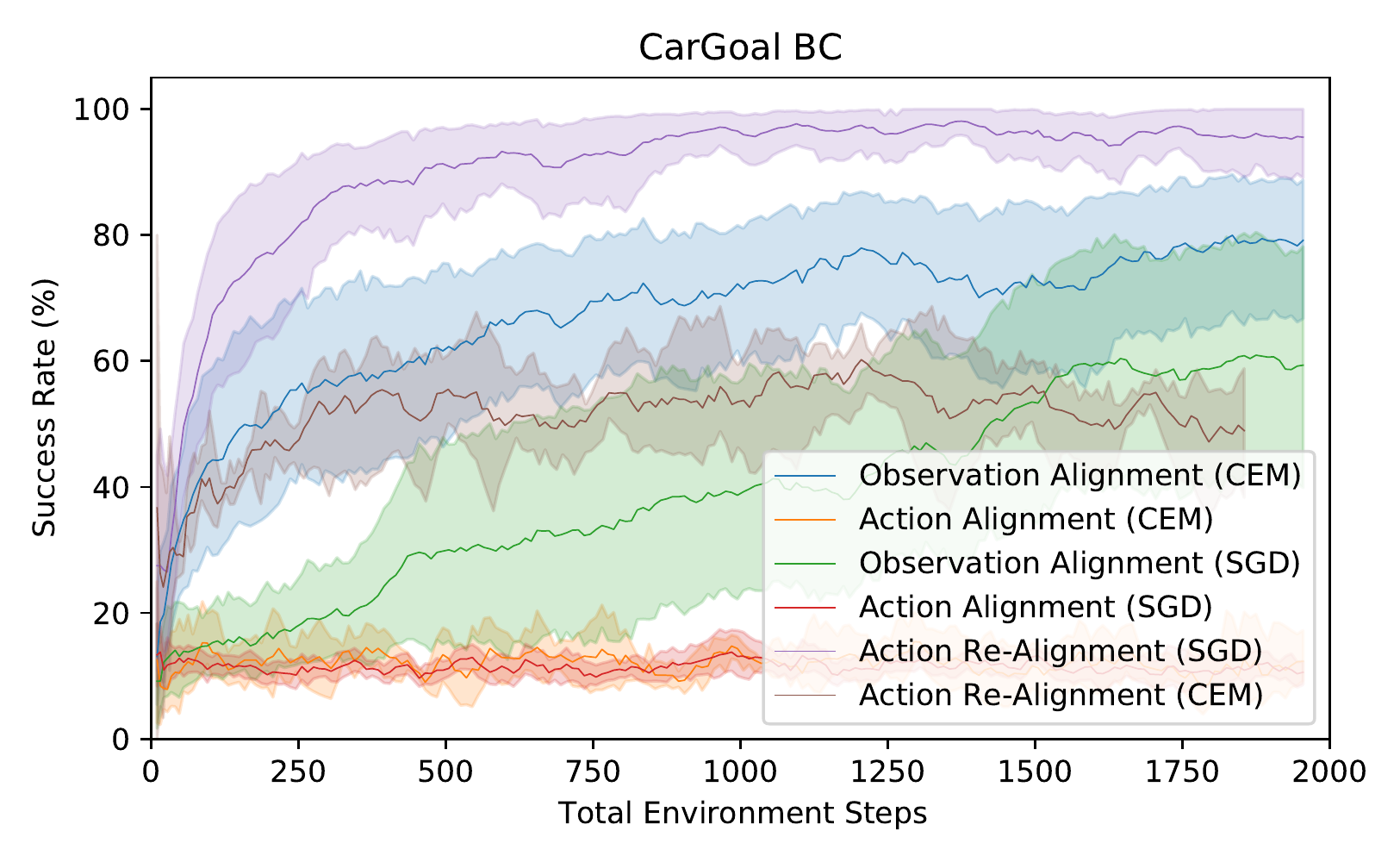}
        \begin{minipage}[t]{0.95\columnwidth}
        \caption{
    This figure shows the performance of different target policy classes and optimization methods using a behavioral cloning loss. The shaded regions represent a 95\% confidence interval. The highest performing methods are Action Re-Alignment with $T_\theta(h)$ trained using SGD, and Observation Alignment with $T_\theta(s)$ trained using CEM.
        This environment takes roughly 200,000 time steps to solve using PPO.
        }
        \label{fig:car_goal_bc}
        \end{minipage}
    \end{minipage}
    \begin{minipage}[t]{0.45\columnwidth}
        \centering
        \includegraphics[width=0.75\columnwidth,valign=t]{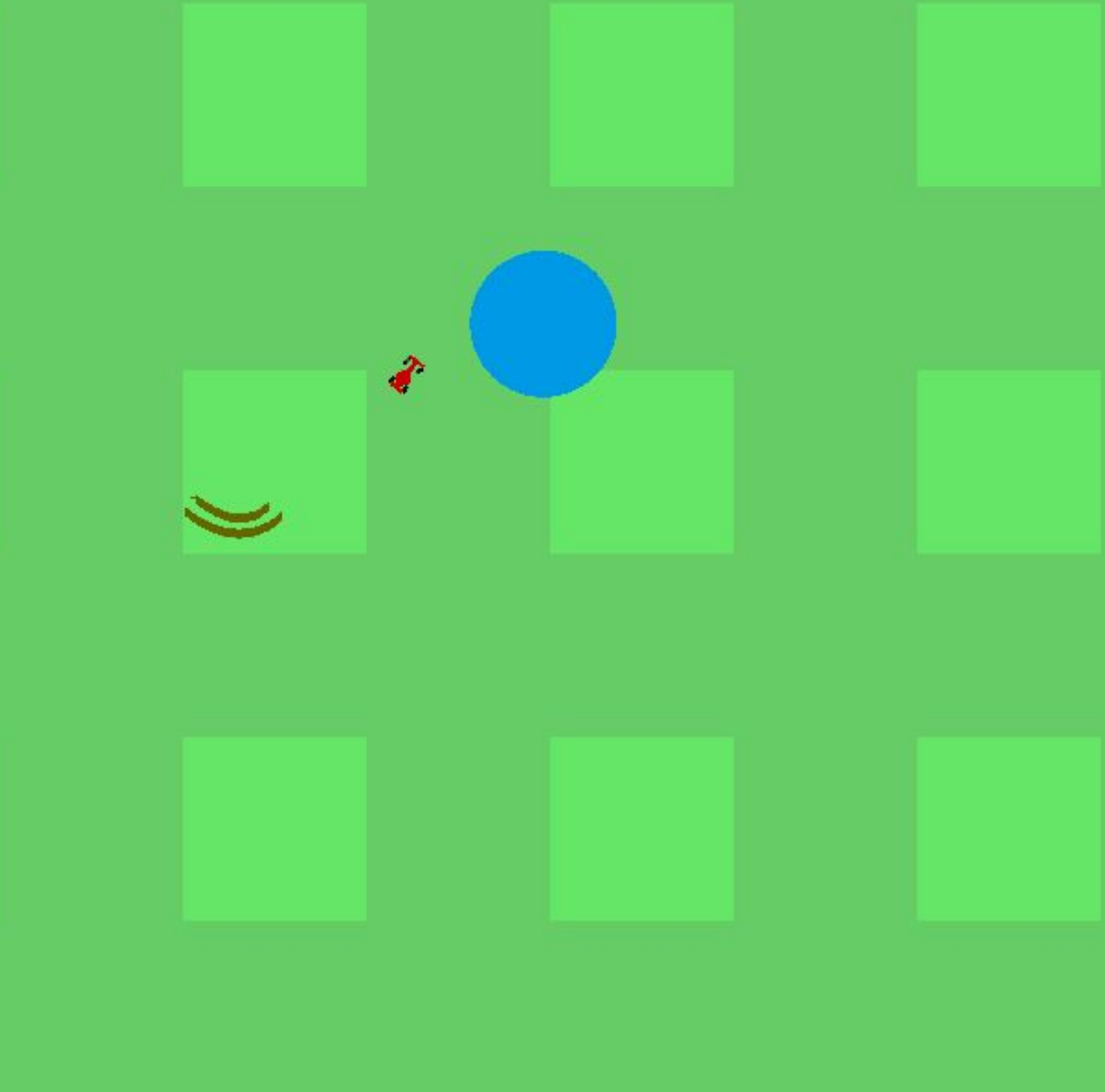}
        \begin{minipage}[t]{0.95\columnwidth}
        \caption{A screenshot of the CarGoal environment. 
        The goal region shown in the image is not observable by the policy.
        }
        \label{fig:car_goal_screenshot}
        \end{minipage}
    \end{minipage}
    \vspace{-0.7cm}
\end{figure}

We find that gradient-based RL and IL methods (BC, AWR, and PPO) can encounter local optima and have difficulty training such low-dimensional function approximators. 
However, we find that the Cross-Entropy Method (CEM) can be used to reliably train the parameters of $T$.
Furthermore, by using CEM with a behavioral cloning loss, instead of Monte-Carlo estimates of expected returns, we are able to re-use a small number of demonstration timesteps to train as success target policy using roughly 2000 timesteps of expert demonstrations in the CarGoal environment.
More details of these results can be seen in Figure~\ref{fig:car_goal_bc}.

\begin{figure}
    \centering
    \begin{minipage}[t]{0.45\columnwidth}
        \centering
        \includegraphics[width=\columnwidth,valign=t]{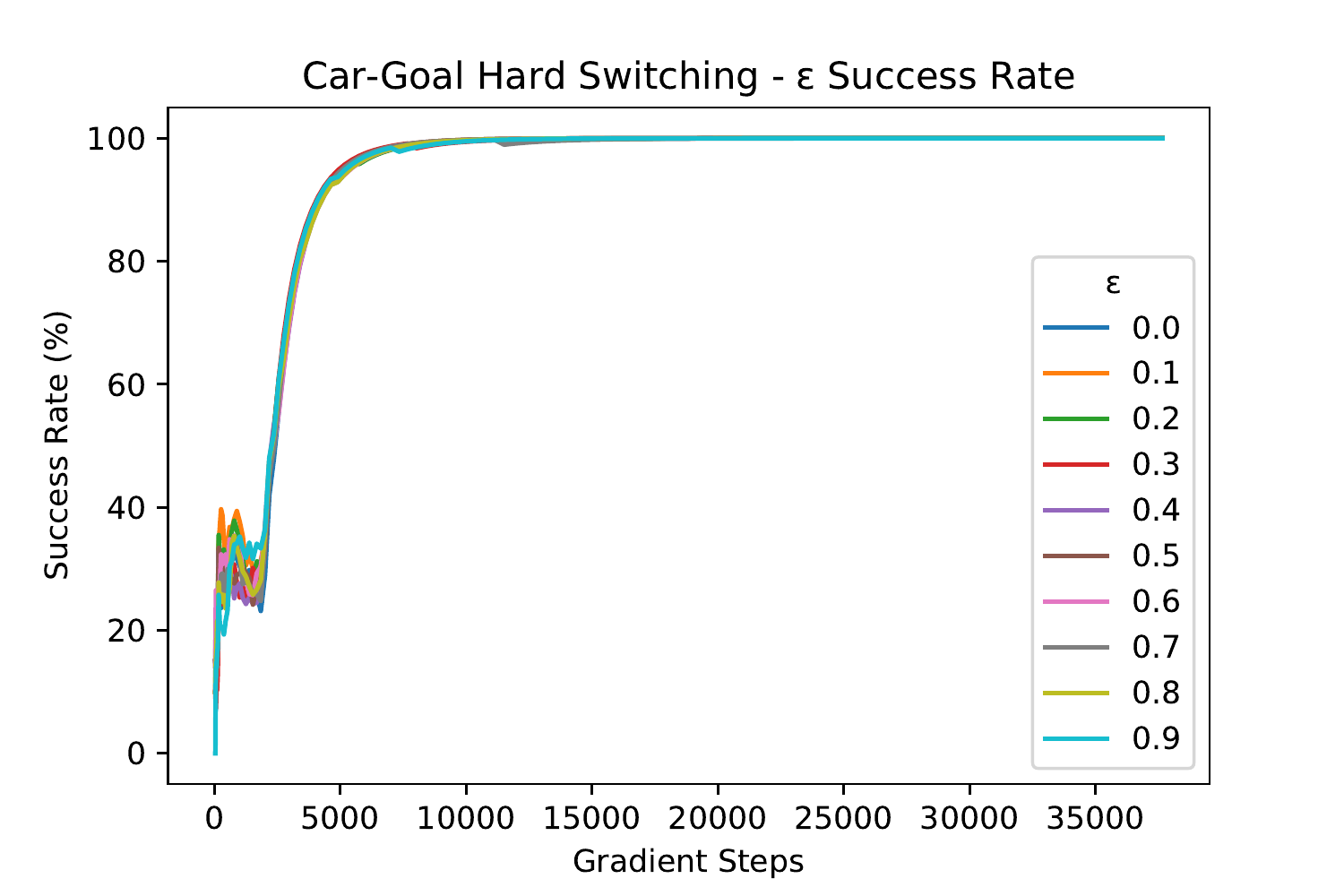}
        \begin{minipage}[t]{0.98\columnwidth}
        \caption{
        This figure shows the shows the performance of hard switching during the training process for a variety of $\epsilon$ values. Note that $\epsilon = 0$ performs equivalently to soft-switching.
        The base policies are three policies trained to reach different goal regions in CarGoal. 
        The target task goal region is outside of the convex hull of those goal regions.
        }
        \label{fig:car_goal_switch_perf}
        \end{minipage}
    \end{minipage}
    \begin{minipage}[t]{0.45\columnwidth}
        \centering
        \includegraphics[width=\columnwidth,valign=t]{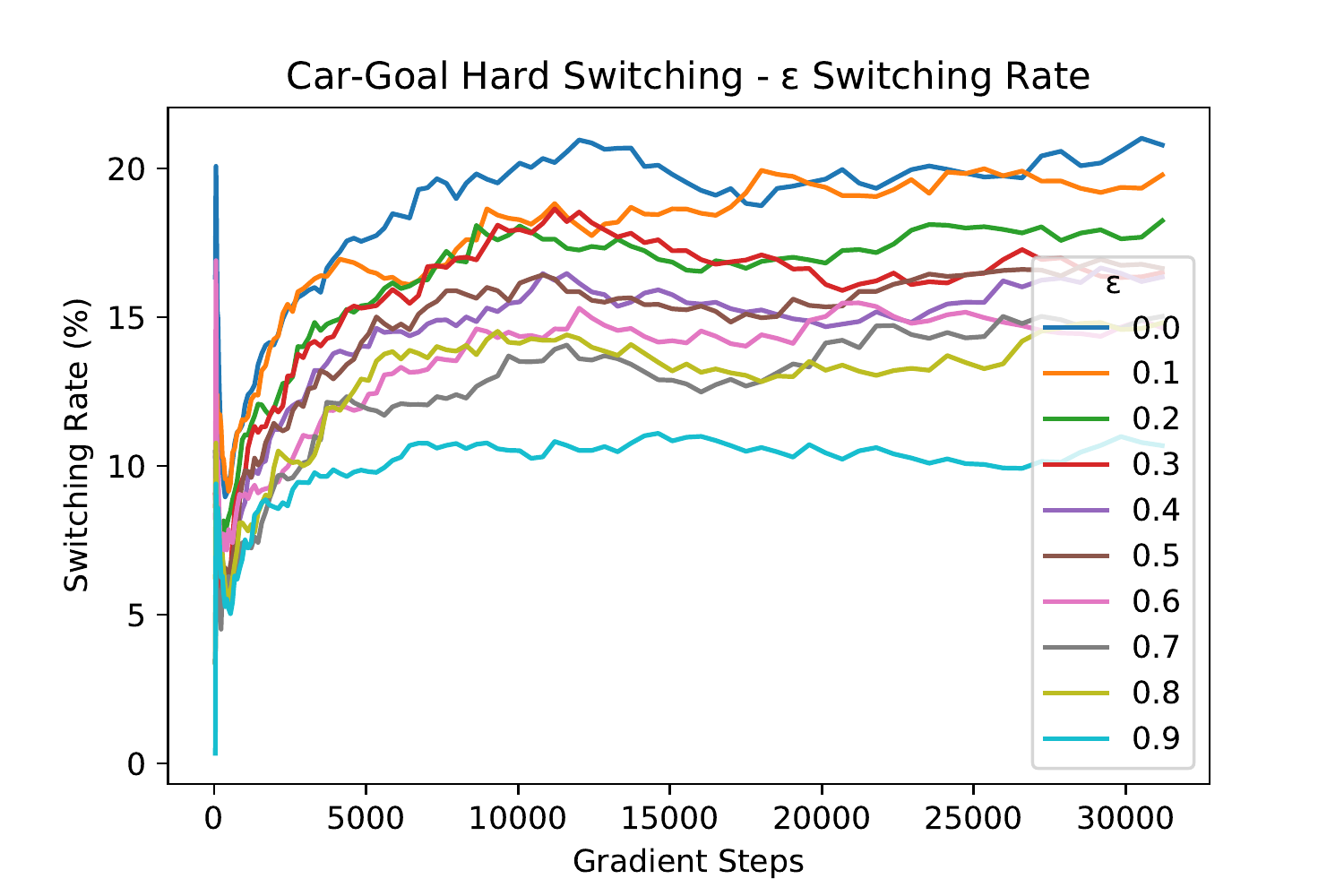}
        \begin{minipage}[t]{0.95\columnwidth}
        \caption{
        This figure shows the average switching rate decreases as epsilon is increased.
        In conjunction with Figure~\ref{fig:car_goal_switch_perf}, it shows that a policy trained using our loss function can switch policies relatively slowly without any visible decrease in the success rate. 
        In this experiment, $\alpha = 0.9$.
        }
        \label{fig:car_goal_hard_switch}
        \end{minipage}
    \end{minipage}
    \vspace{-0.75cm}
\end{figure}


\subsection{Training $W_\theta(s,\tau)$}\snu
In our experiments, we use a fully connected neural network followed by softmax to approximate $W_\theta(s,\tau)$.
We find that it's possible to infer $\theta$ by minimizing a standard behavioral cloning loss (BC) across the target dataset $\mathcal{D}_{\text{target}}$. 
However, we would prefer target policies which switch between base policies less frequently.
To discourage switching, we regularize the standard cross-entropy loss between $W_\theta(s,\tau)$ and the posterior action probabilities $\pi^{\text{base},\tau}(s, a)$, using the cross-entropy loss between $W_\theta(s,\tau)$ and $W_\theta(s',\tau)$, the weights on adjacent states under the target dataset, and combine these terms using a coefficient $\alpha$, which we typically set slightly below $1$.
The effectiveness of this loss function can be seen in Figure~\ref{fig:car_goal_switch_perf} and Figure~\ref{fig:car_goal_hard_switch}.
\[
    L(\theta, s, a, s') = \alpha \mathcal{CE}[W_\theta(s, \tau), \pi^{\text{base},\tau}(a|s))] + (1 - \alpha)\mathcal{CE}[W_\theta(s, \tau), W_\theta(s', \tau)]
\]


\subsection{Environment}\snu
We use a simple goal-conditioned toy environment we call CarGoal, which is based on the CarRacing environment from OpenAI Gym~\cite{brockman2016openai}. The objective of CarGoal is to drive a car to a goal point in the environment, which is hidden from the policy. The policy is rewarded for pointing the car towards the goal, getting the car near the goal, and reaching the goal region.
Reaching the goal region within 1000 time steps is considered ``success,'' and terminates the episode.
Different tasks in this environment correspond to different goal points.
A screenshot of CarGoal is shown in Figure~\ref{fig:car_goal_screenshot}.

\section{Conclusion}\snu
Our results show that the combination of simple architectural approaches, a base set of black-box policies, and simple optimization algorithms like CEM and SGD can produce strong off-distribution transfer results in an environment with non-trivial dynamics. In the future, we look forward to applying these ideas to much more complex problems using real and simulated robotics tasks, and using them to design algorithms for continual robot learning.

\section{Related Work}\snu
Both recent~\cite{van2020mdp}\cite{van2020plannable} and less-recent~\cite{ravindran2001symmetries},\cite{fernandez2006probabilistic} work have explored automatically identifying and exploiting structure, especially symmetries, in general MDPs to speed learning, outside of the transfer learning setting. 
Structural priors, such as symmetry, temporal coherence, and rigid body transformations, have been previously used successfully to adapt deep learning methods to the robotics domain~\cite{jonschkowski2015learning}\cite{byravan2017se3}. Prior works in this area have mostly focused on learning sparse and informative state representations, rather than adapting policies to new tasks using these priors.

Time-domain composition of sub-policies has been studied extensively in hierarchical reinforcement learning, most notably by works using the options framework~\cite{sutton1999between} in which RL sub-policies (options) choose their own termination conditions. 
Other works have studied a problem setting that is more comparable to our own, in which the subpolicies are chosen by a higher-level control process~\cite{gupta2019relay}\cite{fernandez2006probabilistic}. 
Most work in this area has focused on fast transfer by conditioning learned policies on pre-defined goal spaces~\cite{nachum2018data} or grounded representations such as language~\cite{jiang2019language} or known object identities~\cite{zhang2018composable}.

The cross-entropy method~\cite{rubinstein1997optimization} has been used both for direct policy search~\cite{mannor2003cross}, and more recently as an inner search component of value-based RL algorithms, such as to choose actions given a continuous Q-value approximation~\cite{kalashnikov2018scalable}, and to regularize a policy gradient-based action selection algorithm~\cite{shao2020grac}.

Our work is most similar in spirit to Residual Policy Learning~\cite{silver2018residual}, which also uses a pre-trained policy in one task for structured exploration in another task, and uses a deterministic policy target model class of the form $\pi_{\text{target}}(s) = \pi_{\text{base}}(s) + f_\delta(s)$ to facilitate fast adaptation. 

\small
\newpage

\bibliography{references}
\bibliographystyle{unsrt}

\end{document}